\documentclass[10pt,twocolumn,letterpaper]{article}

\usepackage{wacv}              

\usepackage{graphicx}
\usepackage{amsmath}
\usepackage{amssymb}
\usepackage{booktabs}

\usepackage{tabularx}
\usepackage{booktabs}
\usepackage{array}
\usepackage[utf8]{inputenc} 
\usepackage[T1]{fontenc}    
\usepackage{multirow}
\usepackage{url}            
\usepackage{booktabs}       
\usepackage{amsfonts}       
\usepackage{nicefrac}       
\usepackage{microtype}      
\usepackage[table,xcdraw,dvipsnames]{xcolor}         
\usepackage{graphicx}
\usepackage{amsmath}
\usepackage{pgfplots}
\pgfplotsset{compat=1.9}
\usepackage{pgf-pie}
\usepackage{float}
\usepackage{subcaption}
\usepackage[skip=0.333\baselineskip]{caption}
\usepackage[pagebackref,breaklinks,colorlinks]{hyperref}

\usepackage[capitalize]{cleveref}
\crefname{section}{Sec.}{Secs.}
\Crefname{section}{Section}{Sections}
\Crefname{table}{Table}{Tables}
\crefname{table}{Tab.}{Tabs.}

\def\wacvPaperID{1886} 
\def\confName{WACV}
\def\confYear{2025}

\begin{document}

\title{LiGAR: LiDAR-Guided Hierarchical Transformer for \\ Multi-Modal Group Activity Recognition}
\author{Naga Venkata Sai Raviteja Chappa \hspace{10mm} Khoa Luu \\
\\
\emph{CVIU Lab, Department of EECS}\\
\emph{University of Arkansas, Fayetteville USA}\\
{\tt\small \{nchappa,khoaluu\}@uark.edu} \\
\url{https://uark-cviu.github.io/projects/LiGAR/}
}

\maketitle

\begin{abstract}
Group Activity Recognition (GAR) remains challenging in computer vision due to the complex nature of multi-agent interactions. This paper introduces LiGAR, a LIDAR-Guided Hierarchical Transformer for Multi-Modal Group Activity Recognition. LiGAR leverages LiDAR data as a structural backbone to guide the processing of visual and textual information, enabling robust handling of occlusions and complex spatial arrangements. Our framework incorporates a Multi-Scale LIDAR Transformer, Cross-Modal Guided Attention, and an Adaptive Fusion Module to integrate multi-modal data at different semantic levels effectively. LiGAR's hierarchical architecture captures group activities at various granularities, from individual actions to scene-level dynamics. Extensive experiments on the JRDB-PAR, Volleyball, and NBA datasets demonstrate LiGAR's superior performance, achieving state-of-the-art results with improvements of up to 10.6\% in F1-score on JRDB-PAR and 5.9\% in Mean Per Class Accuracy on the NBA dataset. Notably, LiGAR maintains high performance even when LiDAR data is unavailable during inference, showcasing its adaptability. Our ablation studies highlight the significant contributions of each component and the effectiveness of our multi-modal, multi-scale approach in advancing the field of group activity recognition.
\end{abstract}

\section{Introduction}\label{sec:intro}

\begin{figure}
    \centering
    \includegraphics[width=\linewidth]{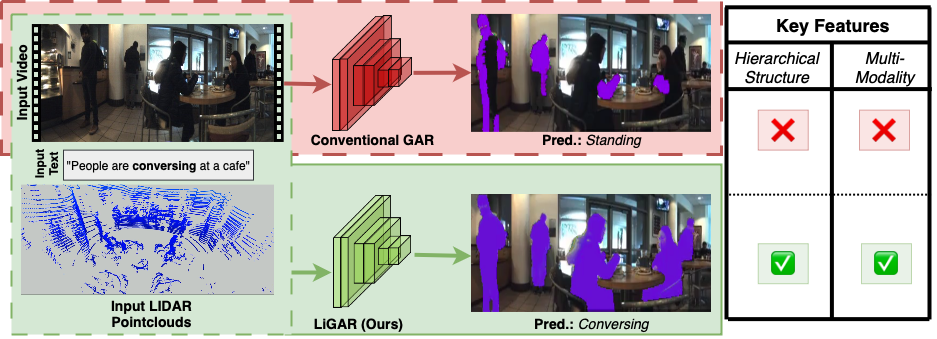}
    \caption{Comparison of LiGAR with conventional GAR methods~\cite{chappa2023spartan, han2022dual} in analyzing group activities. LiGAR leverages multi-modal data (video, text, and LiDAR point clouds) to provide a more accurate scene classification, capturing the nuanced group activity of 'Conversing.' The key features highlighted show LiGAR's superior performance in hierarchical structure understanding and multi-modal integration compared to previous methods. \textbf{Best viewed in color and zoom.}}
    \label{fig:motivation}
\end{figure}

Group Activity Recognition (GAR) has emerged as a crucial challenge in computer vision, bridging the gap between individual action recognition and complex scene understanding. Recent years have witnessed significant advancements in this field, with approaches ranging from hierarchical models~\cite{ibrahim2016hierarchical} and actor relation graphs~\cite{wu2019learning} to transformer-based architectures~\cite{gavrilyuk2020actor, li2021groupformer}. However, these methods often interpret group activities in a constrained manner, typically relying on a single modality or processing multiple modalities independently before fusion. This approach limits the potential for discovering more nuanced and diverse group interactions. It is well illustrated in~\cref{fig:motivation}.

Delving into the multi-modal GAR problem, we introduce LiDAR-Guided Hierarchical Transformer (LiGAR), a novel framework that fundamentally reimagines the approach to GAR. LiGAR is characterized by its ability to process and integrate information from multiple modalities LiDAR, video, and text at multiple scales, capturing a rich spectrum of spatial, temporal, and semantic information. To this end, we propose a multi-scale processing pipeline that forms the backbone of LiGAR, offering an intuitive modeling technique for enriched comprehension of group activities and complex inter-actor dynamics.

The proposed LiGAR framework operates with a unique \textit{Cross-Modal Guided Attention} mechanism at every level to jointly process multi-modal information. This strategy simplifies operations and eliminates the intricacies of separate modality-specific processing streams. Instead of perceiving group activities as a monolithic block, LiGAR models the input data with a \textit{hierarchical structure}, promoting a holistic grasp of actor interplays. Each level delves into essential insights, benefiting from the features of different scales and modalities, capturing scene changes more comprehensively over time.

In addition, LiGAR promotes dynamic \textit{adaptability} and \textit{flexibility} through its adaptive modality fusion module, empowering the model to adjust its focus on different modalities to capture group activities throughout video sequences. This adaptability is further showcased as LiGAR proficiently tackles group activity recognition across diverse scenarios, from sports analytics to surveillance systems, demonstrating its extensive flexibility in decoding various interaction nuances. The proposed framework is not confined to specific domains, emphasizing its broad applicability and potential.

\textbf{The Contributions of this Work.} We present our critical contributions in a threefold manner as below:

1) \textbf{LiDAR-Guided Multi-Modal Architecture:} We propose a novel hierarchical transformer architecture that leverages LiDAR data as a structural backbone to guide the processing of visual and textual information. This includes a Multi-Scale LiDAR Transformer (MLT), as presented in~\cref{sec:mlt}, for creating hierarchical scene representations and a Cross-Modal Guided Attention (CMGA) mechanism, presented in~\cref{sec:cmga}. This innovative approach enables a more robust handling of complex spatial arrangements and occlusions, significantly enhancing the model's ability to interpret nuanced group activities.

2) \textbf{Adaptive Multi-Modal Fusion and Hierarchical Decoding:} We introduce an Adaptive Fusion Module (AFM), presented in~\cref{sec:afm}, based on TimeSformer that dynamically integrates LiDAR, visual, and textual modalities while modeling temporal dependencies. This is coupled with a Hierarchical Activity Decoder, described in~\cref{sec:decoder}, that predicts activities at multiple levels of granularity, from individual actions to scene-level dynamics. These components allow LiGAR to adaptively focus on different modalities and scales throughout video sequences, capturing the full spectrum of group interactions.

3) \textbf{Comprehensive Evaluation:} Through extensive experiments on diverse benchmark datasets, including JRDB-PAR~\cite{han2022panoramic}, Volleyball~\cite{ibrahim2016hierarchical}, and NBA~\cite{yan2020social}, we demonstrate LiGAR's superiority in multi-modal group activity recognition. Our results show significant improvements over state-of-the-art methods, with gains of up to 10.6\% in F1-score on JRDB-PAR and 5.9\% in Mean Per Class Accuracy on the NBA dataset. Notably, LiGAR maintains high performance even when LiDAR data is unavailable during inference, showcasing its adaptability and potential for broad impact across different application domains.

Our work advances the state-of-the-art in GAR and opens new avenues for leveraging multimodal, multiscale information in various computer vision tasks. The LiGAR framework's ability to perform effectively even when LiDAR data is unavailable during inference highlights its adaptability and potential for broad impact across different application domains.

\begin{figure*}[ht]
\centering
\includegraphics[width=\textwidth]{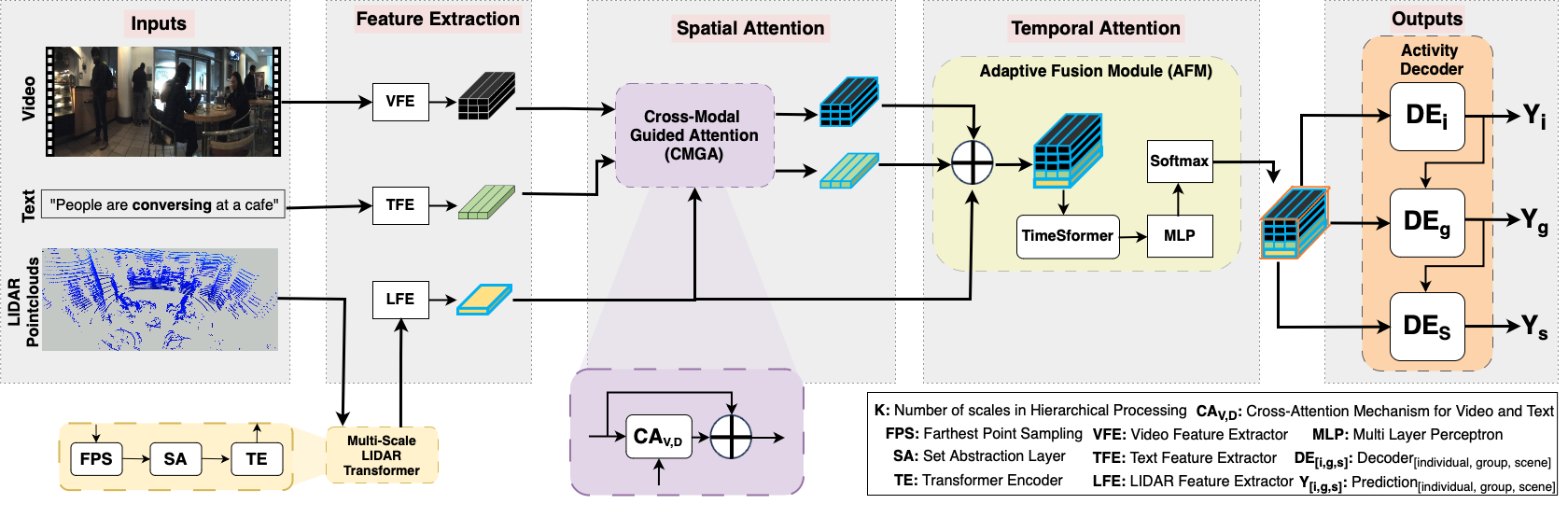}
\put(-489,145){\rotatebox{90}{\tiny($\mathbf{V}$)}}
\put(-489,109){\rotatebox{90}{\tiny($\mathbf{D}$)}}
\put(-489,88){\rotatebox{90}{\tiny($\mathbf{L}$)}}
\put(-367,41){\tiny$\mathbf{Z_k}$}
\put(-339,65){\tiny$\mathbf{F_L^k}$}
\put(-339,95){\tiny$\mathbf{F_D^k}$}
\put(-339,125){\tiny$\mathbf{F_V^k}$}
\put(-314,22){\tiny$\mathbf{F_{V,D}^k}$}
\put(-286,16){\tiny$\mathbf{F_L^k}$}
\put(-225,146){\tiny$\mathbf{V'^k}$}
\put(-225,106){\tiny$\mathbf{D'^k}$}
\put(-146,129){\tiny$\mathbf{X_t'^k}$}
\put(-91,79){\tiny$\mathbf{F_{fused}}$}
\caption{Overview of the LiDAR-Guided Hierarchical Transformer (LiGAR) framework for multi-modal group activity recognition.  It effectively integrates multi-modal inputs, including Video, Text, and LIDAR, through specialized feature extractors (VFE, TFE, and Multi-Scale LIDAR Transformer with LFE). The Cross-Modal Guided Attention (CMGA) mechanism aligns features across modalities, detailed in~\cref{sec:cmga}. The Adaptive Fusion Module (AFM) with TimeSformer~\cite{bertasius2021space} dynamically weights modal contributions for temporal modeling, as explained in~\cref{sec:afm}. Finally, the Hierarchical Activity Decoder ($DE_i$, $DE_g$, $DE_s$) produces predictions at individual, group, and scene levels, capturing the multi-granular nature of group activities, as described in~\cref{sec:decoder}. \textbf{Best viewed in zoom and color.}}

\label{fig:LiGAR_framework}
\end{figure*}
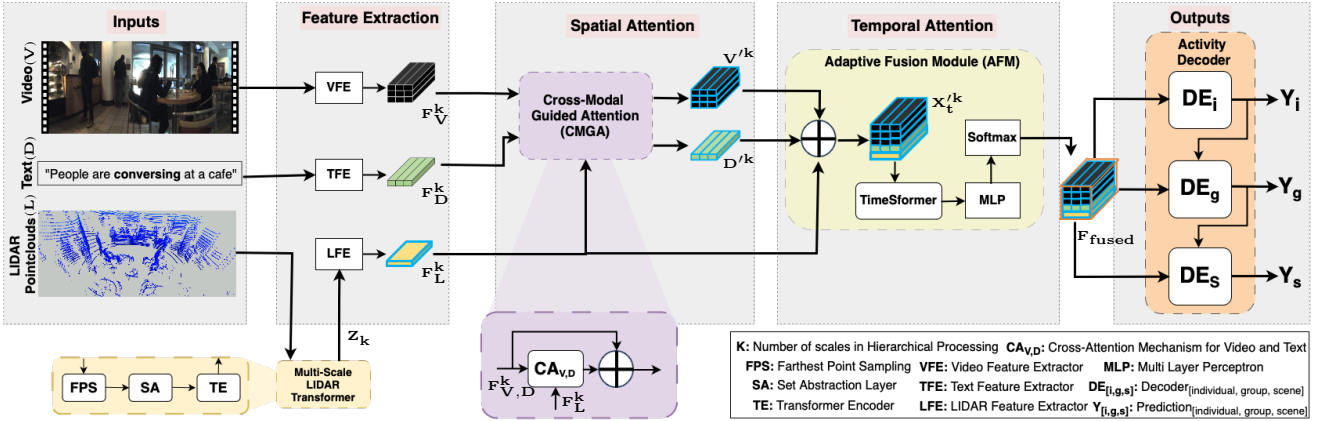
\section{Related Work} \label{sec:related_works}

\subsection{Group Activity Recognition}

Group Activity Recognition has emerged as a significant area of research in computer vision, with applications spanning various domains~\cite{bagautdinov2017social, deng2016structure, ibrahim2016hierarchical, ibrahim2018hierarchical, li2017sbgar, qi2018stagnet, shu2019hierarchical, wang2017recurrent, yan2018participation, chappa2023sogar, chappa2023spartan}. The typical GAR paradigm involves analyzing video clips to predict group activity labels, often requiring an understanding of individual actions and their collective implications.

Early GAR methods relied on hand-engineered features and traditional machine learning models like random forests. The advent of deep learning transformed the field, introducing methods that leverage multi-level RNN architectures to capture both individual and interactive features.

Recent advancements in GAR have predominantly focused on relation learning, employing sophisticated deep learning techniques. Graph neural networks and attention-based architectures~\cite{han2022dual, gavrilyuk2020actor} have become prevalent, enabling more nuanced modeling of inter-actor relationships. These approaches often involve extracting human-level features using techniques like RoIPooling~\cite{qi2018stagnet} or RoIAlign~\cite{tamura2022hunting}, followed by relation learning on these extracted features.

\subsection{Hierarchical and Multi-modal Approaches in GAR}

Recognizing the complex nature of group activities, researchers have explored both hierarchical and multi-modal approaches to enhance GAR performance. Hierarchical methods aim to capture activities at various semantic levels, while multi-modal approaches leverage diverse data sources for richer representations.

Several works have explored hierarchical structures in GAR. \cite{ibrahim2016hierarchical} proposed a hierarchical LSTM model to capture activities at individual, group, and scene levels. \cite{ibrahim2018hierarchical} introduced a multi-scale temporal CNN to process activities at various temporal resolutions. More recently, transformer-based architectures have been employed for multi-level activity recognition. \cite{chappa2024hatt} utilized a hierarchical transformer to model long-range dependencies across different semantic levels.

Concurrently, multi-modal approaches have gained traction in addressing the limitations of single-modality methods. \cite{Shahroudy_2016_CVPR} combined RGB videos with depth information to improve activity recognition in complex scenarios. Of particular relevance to our work, several recent studies have explored the integration of LiDAR data in activity recognition tasks. Sun~\etal~\cite{sun2024x} used LiDAR point clouds to enhance 3D human pose estimation for improved action recognition, while Jiao~\etal~\cite{Jiao_2023_CVPR} proposed a fusion network that combines features from RGB and LiDAR modalities for robust activity classification. Chappa~\etal~\cite{chappa2023react} introduced a multi-modal framework using RGB and text inputs to perform group activity recognition and grounding actions in a scene.

Despite these advancements, the integration of hierarchical modeling with truly multimodal approaches remains largely unexplored. Many existing methods either focus on hierarchical structures within a single modality or employ late fusion strategies for multimodal data, potentially missing critical intermodal relationships at different semantic levels. Additionally, incorporating textual data alongside visual and LiDAR information in a hierarchical framework for GAR remains an open challenge.

\section{Methodology}\label{sec:methodology}

\subsection{Overview}

The LiGAR framework, as illustrated in~\cref{fig:LiGAR_framework}, introduces a novel approach to multi-modal group activity recognition by leveraging the strengths of LIDAR, video, and textual data. Our method processes information at multiple scales throughout the pipeline, enabling a rich understanding of group activities at various levels of granularity.

The key innovations of our approach include: 1.) Multi-scale processing of LIDAR data to capture spatial information at different resolutions. 2.) Cross-modal guided attention mechanism that uses LIDAR features to enhance video and text representations. 3.) Adaptive fusion of modalities using TimeSformer, allowing for dynamic weighting of information sources. 4.) A hierarchical decoder that predicts activities at the scene, group, and individual levels.

The following subsections detail each framework component, explaining the intuition behind our design choices and how they contribute to effective group activity recognition.

\subsection{Multi-Scale LiDAR Processing}\label{sec:mlt}

LIDAR data provides valuable 3D spatial information that can significantly enhance our understanding of group activities. To fully leverage this information, we process LIDAR point clouds at multiple scales, capturing fine-grained details and broader spatial contexts.

\subsubsection{Multi-Scale LiDAR Transformer (MLT)}

The MLT processes point clouds at $K$ different scales, creating a hierarchical scene representation. For each scale $k$, we perform the following operations:
\begin{equation}
    H_k = \text{SA}_k(\text{FPS}_k(L)), \quad k \in \{1, ..., K\}
\end{equation}
\begin{equation}
    Z_k = \text{TransformerEncoder}_k(H_k)
\end{equation}
Here, $\text{FPS}_k$ is the farthest point sampling operation, which selects a subset of points to represent the scene at scale $k$. $\text{SA}_k$ is a set abstraction layer from PointNet++~\cite{qi2017pointnet++}, which aggregates local features. The TransformerEncoder then processes these features, capturing complex spatial relationships.

This multi-scale approach allows our model to capture fine-grained interactions and broader spatial arrangements (e.g., group formations) simultaneously.

\subsubsection{LiDAR Feature Extractor (LFE)}

To further refine the LIDAR representations for activity recognition, we introduce a LiDAR Feature Extractor:
\begin{equation}
    F_L^k = \text{LFE}_k(Z_k), \quad k \in \{1, ..., K\}
\end{equation}
The LFE uses scale-specific attention mechanisms to focus on spatiotemporal patterns indicative of group activities at different granularity levels. It allows our model to learn features relevant to activity recognition from the rich LIDAR data.

\subsection{Multi-modal Feature Extraction}

While LIDAR data provides valuable spatial information, video and text modalities offer complementary cues crucial for comprehensive activity recognition. We process these modalities at multiple scales to maintain consistency with our LIDAR processing pipeline.

\subsubsection{Video Feature Extractor (VFE)}

For video data, we employ a pyramid of 3D ConvNets followed by scale-specific temporal attention mechanisms:
\begin{equation}
    F_V^k = \text{VFE}_k(V), \quad k \in \{1, ..., K\}
\end{equation}
This multi-scale approach lets us capture fine-grained motion patterns and broader temporal contexts in the video data.

\subsubsection{Text Feature Extractor (TFE)}

To process textual descriptions, we use a multi-level language model that captures both fine-grained semantic information and high-level concepts:
\begin{equation}
    F_D^k = \text{TFE}_k(D), \quad k \in \{1, ..., K\}
\end{equation}
This hierarchical text processing allows our model to understand specific actions described in the text and overarching themes or contexts.

\subsection{Cross-Modal Guided Attention (CMGA)}\label{sec:cmga}

A key innovation in our framework is using LIDAR information to guide video and text data processing. This cross-modal attention mechanism allows us to leverage the precise spatial information from LIDAR to enhance our understanding of the visual and textual inputs.

For the visual domain at each scale $k$:
\begin{equation}
    A_V^k = \text{softmax}(\frac{(W_Q^k F_V^k)(W_K^k F_L^k)^T}{\sqrt{d_k}})
\end{equation}
\begin{equation}
    V'^k = A_V^k (W_V^k F_L^k) + F_V^k
\end{equation}
Similarly, for the textual domain:

\begin{equation}
    A_D^k = \text{softmax}(\frac{(W_Q^k F_D^k)(W_K^k F_L^k)^T}{\sqrt{d_k}})
\end{equation}
\begin{equation}
    D'^k = A_D^k (W_V^k F_L^k) + F_D^k
\end{equation}
This mechanism allows the model to focus on the most relevant parts of the video or text, given the spatial configuration captured by the LIDAR data. For example, it might help the model pay more attention to visual features in areas where the LIDAR data indicates the presence of people.

\subsection{Adaptive Fusion Module (AFM)}\label{sec:afm}

To effectively combine information from all three modalities (LiDAR, visual, and depth), we introduce an Adaptive Fusion Module (AFM) based on TimeSformer~\cite{bertasius2021space}. It allows our model to dynamically adjust the importance of each modality based on the current context, while also modeling temporal dependencies across frames.
Our AFM operates at multiple scales to capture both fine-grained and coarse features. For each scale $k$, we first concatenate the features from all modalities:
\begin{equation}
X_t^k = [F_L^k; V'^k; D'^k], \quad k \in {1, ..., K}
\end{equation}
where $F_L^k$ represents LiDAR features, $V'^k$ represents visual features, and $D'^k$ represents depth features at scale $k$. The concatenated feature $X_t^k$ combines information from all modalities for the current time step $t$.
We then process this concatenated feature using TimeSformer, a transformer-based architecture designed for video understanding:
\begin{equation}
Z_t^k = \text{TimeSformer}k(X{1:t}^k), \quad k \in {1, ..., K}
\end{equation}
TimeSformer applies spatially and temporally self-attention mechanisms, allowing it to capture complex spatiotemporal relationships in the multimodal data. The output $Z_t^k$ encodes these relationships for each scale.
The output of TimeSformer is then used to compute adaptive weights for each modality:
\begin{equation}
\alpha_t^k = \text{Softmax}(\text{MLP}_k(Z_t^k)), \quad k \in {1, ..., K}
\end{equation}
Here, $\text{MLP}_k$ is a multi-layer perceptron that maps the TimeSformer output to a three-dimensional vector (one dimension per modality). The softmax function ensures that the weights sum to 1, allowing us to interpret them as the relative importance of each modality.
These weights are then used to fuse the modalities:
\begin{equation}
F_{\text{fused}}^k = \alpha_t^k[0] \cdot F_L^k + \alpha_t^k[1] \cdot V'^k + \alpha_t^k[2] \cdot D'^k
\end{equation}
This weighted sum allows the model to dynamically emphasize the most informative modality for each particular instance and scale.
The final fused representation combines information from all scales:
\begin{equation}
F_{\text{fused}} = [F_{\text{fused}}^1; F_{\text{fused}}^2; ...; F_{\text{fused}}^K]
\end{equation}
By concatenating the fused features from all scales, we obtain a multi-scale representation that captures both fine-grained details and higher-level structure.
This adaptive fusion mechanism enhances our model's robustness across diverse scenarios. For instance, in low-light conditions, the model might rely more heavily on LiDAR and depth information, while in complex urban environments with many visual cues, it might prioritize visual features. The multi-scale nature of the fusion also allows the model to adaptively focus on the most relevant spatial scales for each task and environment.

\subsection{Hierarchical Activity Decoder}\label{sec:decoder}

Our hierarchical activity decoder predicts activities at multiple levels of granularity, reflecting the hierarchical nature of group activities. We use a cascaded approach, where predictions at broader levels inform those at finer levels:
\begin{equation}
    Y_{\text{individual}} = \text{Decoder}_{\text{individual}}(F_{\text{fused}})
\end{equation}
\begin{equation}
    Y_{\text{group}} = \text{Decoder}_{\text{group}}(F_{\text{fused}}, Y_{\text{individual}})
\end{equation}
\begin{equation}
    Y_{\text{scene}} = \text{Decoder}_{\text{scene}}(F_{\text{fused}}, Y_{\text{group}})
\end{equation}
This hierarchical decoding allows our model to maintain consistency across different levels of prediction and leverage higher-level context for fine-grained predictions.

\subsection{Loss Function}

Our loss function is designed to encourage accurate predictions at all scales and levels of granularity:
\begin{equation}
    \mathcal{L} = \sum_{k=1}^K w_k \mathcal{L}_{\text{task}}^k + \lambda_1 \mathcal{L}_{\text{HCL}} + \lambda_2 \mathcal{L}_{\text{temporal}} 
\end{equation}
Here, $\mathcal{L}_{\text{task}}^k$ is the task loss at scale $k$, which includes losses for scene, group, and individual level predictions. The weights $w_k$ allow us to balance the importance of different scales.

$\mathcal{L}_{\text{HCL}}$ is a Hierarchical Consistency Loss that encourages consistency between predictions at different levels (e.g., ensuring that individual actions are compatible with the predicted group activity).

$\mathcal{L}_{\text{temporal}}$ is a temporal consistency loss that encourages smooth and coherent predictions across frames.

Combining these loss terms ensures that our model learns to make accurate, consistent, and temporally coherent predictions at all granularity levels.

\subsection{Inference}

During inference, our model processes the multi-modal input through the entire pipeline, producing predictions at scene, group, and individual levels. The multi-scale nature of our approach allows for both top-down (using broader context to inform fine-grained predictions) and bottom-up (aggregating fine-grained information to inform higher-level predictions) reasoning about group activities.

This comprehensive approach enables our model to capture the complex, hierarchical nature of group activities, leading to more accurate and interpretable predictions.

\section{Experimental Results}

\subsection{Datasets and Evaluation Metrics}

\noindent\textbf{JRDB-PAR Dataset}~\cite{han2022panoramic}: Contains 27 categories of individual actions, 11 social group activities, and seven global activities. It consists of 27 videos (20 for training, 7 for testing), totaling 27,920 frames. Evaluation uses precision, recall, and F1-score ($\mathcal{P}_g$, $\mathcal{R}_g$, $\mathcal{F}_g$) for multi-label classification of social group activities.

\noindent\textbf{Volleyball Dataset}~\cite{ibrahim2016hierarchical}: Comprises 4,830 labeled clips (3,493 for training, 1,337 for testing) from 55 videos. We focus only on group activity labels for WSGAR (Weakly-Supervised GAR) experiments. Evaluation metrics include Multi-class Classification Accuracy (MCA) and Merged MCA.

\noindent\textbf{NBA Dataset}~\cite{yan2020social}: Contains 9,172 labeled clips (7,624 for training, 1,548 for testing) from 181 NBA videos, annotated only for group activities. The evaluation addresses the class imbalance by using Multi-class Classification Accuracy (MCA) and Mean Per Class Accuracy (MPCA).

\subsection{Implementation Details}
LiGAR is implemented using PyTorch~\cite{paszke2019pytorch} and Open3D 0.13~\cite{zhou2018open3d} for efficient LiDAR point cloud processing. 
For the Video Feature Extractor (VFE), we utilize a 3D ResNet-50~\cite{tran2018closer} architecture, pre-trained on Kinetics-400~\cite{kay2017kinetics}. The Text Feature Extractor (TFE) is based on BERT-base~\cite{devlin2018bert}, which we fine-tune during training.
Our multi-scale processing is implemented with K=3 scales, capturing fine, medium, and coarse-grained features across all modalities. The TimeSformer in our AFM (as mentioned in~\cref{sec:afm}) uses a similar configuration to the original paper~\cite{bertasius2021space}, but with adjustments to handle multi-scale, multi-modal input.

Training strategy:
1. Pre-train LiGAR on JRDB-PAR using all modalities (RGB, LiDAR, IMU).
2. Fine-tune on Volleyball and NBA datasets using only RGB data.
3. For inference on all datasets, we use only RGB video input to demonstrate the model's adaptability.

We use Adam optimizer~\cite{kingma2014adam} with a learning rate of 1e-4 and a batch size of 32. The learning rate is decreased by 0.1 every 20 epochs. Training is performed on 4 NVIDIA A100 GPUs.

\subsection{Comparison with State-of-the-Art Methods}

\noindent\textbf{Performance on JRDB-PAR Dataset:}
We evaluate LiGAR on the challenging JRDB-PAR dataset~\cite{han2022panoramic}, which features complex group activities in diverse real-world scenarios.~\Cref{tab:jrdb-par-results} presents a comparison of our method with several state-of-the-art approaches.

\begin{table}[h]
\centering
\caption{Comparison with SOTA on JRDB-PAR dataset}\label{tab:jrdb-par-results}
\begin{tabular}{lccc}
\hline
\textbf{Method} & $\mathcal{P}_g$ & $\mathcal{R}_g$ & $\mathcal{F}_g$ \\
\hline
ARG~\cite{wu2019learning} & 34.6 & 29.3 & 30.7 \\
SA-GAT~\cite{ehsanpour2020joint} & 36.7 & 29.9 & 31.4 \\
JRDB-Base~\cite{ehsanpour2022jrdb} & 44.6 & 46.8 & 45.1 \\
SACRF~\cite{pramono2020empowering} & 42.9 & 35.5 & 37.6 \\
HiGCIN~\cite{yan2020higcin} & 39.3 & 30.1 & 33.1 \\
SoGAR~\cite{chappa2023sogar} & 49.3 & 47.1 & 48.7 \\
\textbf{LiGAR (Ours)} & \textbf{61.2} & \textbf{58.8} & \textbf{59.3} \\
\hline
\end{tabular}
\end{table}

LiGAR significantly outperforms existing methods across all metrics, achieving 61.2\% precision ($\mathcal{P}_g$), 58.8\% recall ($\mathcal{R}_g$), and 59.3\% F1-score ($\mathcal{F}_g$) improvements of 11.9\%, 11.7\%, and 10.6\% respectively over the previous best method, SoGAR~\cite{chappa2023sogar}. This superior performance stems from LiGAR's effective integration of RGB, LiDAR, and textual data, coupled with its efficient architecture. LiGAR achieves these results, underscoring the effectiveness of our Multi-Scale LiDAR Transformer and Cross-Modal Guided Attention components. These advancements enable LiGAR to address better the complexities of group activity recognition in real-world scenarios, thereby pushing forward the state-of-the-art in multi-modal GAR.

\noindent\textbf{Performance Comparison on Volleyball and NBA Datasets:}~\cref{tab:volleyball_nba_comparison} presents LiGAR's performance against state-of-the-art methods on the Volleyball~\cite{ibrahim2016hierarchical} and NBA~\cite{yan2020social} datasets. In weakly supervised settings, LiGAR consistently outperforms existing methods across both datasets. On the Volleyball dataset, LiGAR achieves 93.1\% MCA and 95.4\% M-MCA, surpassing the previous best weakly supervised method by 1.3\% and 0.9\%, respectively, and even matching or exceeding several fully supervised approaches. For the NBA dataset, LiGAR's performance is particularly notable, with 87.4\% MCA and 79.4\% MPCA, representing significant improvements of 4.1\% and 5.9\% over the previous best. Importantly, LiGAR achieves these results with only 14.2M parameters, making it the most efficient model among its competitors.

The superior performance of LiGAR, especially on the more challenging NBA dataset, can be attributed to its effective integration of multi-modal information and multi-scale processing. This approach allows LiGAR to capture complex spatio-temporal relationships and group dynamics that may be missed by single-modality or single-scale methods. The substantial improvement in MPCA on the NBA dataset (5.9\%) suggests that LiGAR is particularly adept at recognizing a diverse range of group activities, addressing the class imbalance issue common in real-world scenarios. Furthermore, LiGAR's ability to outperform fully supervised methods in a weakly supervised setting demonstrates its robustness to limited annotations. It is a critical advantage for practical applications where detailed labeling may be infeasible. We also performed additional experiments with different backbones in \textit{Sec. 1} of supplementary material.

\begin{table*}[h]
\centering
\caption{Comparison with SOTA on Volleyball and NBA datasets. Here, \textbf{WS}: Weakly Supervised.}\label{tab:volleyball_nba_comparison}
\small
\begin{tabular}{lccccccc}
\hline
\multirow{2}{*}{\textbf{Method}} & \multirow{2}{*}{\textbf{WS}} & \multirow{2}{*}{\textbf{Backbone}} & \multicolumn{2}{c}{\textbf{Volleyball}} & \multicolumn{3}{c}{\textbf{NBA}} \\
\cline{4-8}
 & & & \emph{MCA} & \emph{M-MCA} & \emph{Params} & \emph{MCA} & \emph{MPCA} \\
\hline
StagNet~\cite{qi2018stagnet} & No & VGG-16 & 89.3 & - & - & - & - \\
ARG~\cite{wu2019learning} & No & ResNet-18 & 91.1 & 95.1 & - & - & - \\
HIGCIN~\cite{yan2020higcin} & No & ResNet-18 & 91.4 & - & - & - & - \\
DIN~\cite{yuan2021learning} & No & ResNet-18 & 93.1 & \textbf{95.6} & - & - & - \\
GroupFormer~\cite{li2021groupformer} & No & Inception-v3 & 94.1 & - & - & - & - \\
Dual-AI~\cite{han2022dual} & No & Inception-v3 & \textbf{94.4} & - & - & - & - \\
\hline
ARG~\cite{wu2019learning} & Yes & ResNet-18 & 80.5 & 90.0 & 49.5M & 59.0 & 56.8 \\
AT~\cite{gavrilyuk2020actor} & Yes & ResNet-18 & 84.3 & 89.6 & 29.6M & 47.1 & 41.5 \\
DIN~\cite{yuan2021learning} & Yes & ResNet-18 & 86.5 & 93.1 & 26.0M & 61.6 & 56.0 \\
SAM~\cite{yan2020social} & Yes & Inception-v3 & - & 94.0 & 25.5M & 54.3 & 51.5 \\
DFWSGAR~\cite{kim2022detector} & Yes & ResNet-18 & 90.5 & 94.4 & 17.5M & 75.8 & 71.2 \\
SOGAR~\cite{chappa2023sogar} & Yes & ResNet-18 & 91.8 & 94.5 & 15.1M & 83.3 & 73.5 \\
LiGAR (Ours) & Yes & ResNet-18 & \textbf{93.1} & \textbf{95.4} & \textbf{14.2M} & \textbf{87.4} & \textbf{79.4} \\
\hline
\end{tabular}
\end{table*}

\subsection{Ablation Studies}

We conduct several ablation studies to validate the design choices in LiGAR using the JRDB-PAR dataset; however, we present more ablation study results on Volleyball and NBA datasets, along with the impact of losses, in the Supplementary Material.

\subsubsection{Impact of Multi-modal Fusion}

\begin{table}[h]
\centering
\caption{Impact of modality combinations}
\label{tab:modality_ablation}
\begin{tabular}{lccc}
\hline
\textbf{Modalities} & $\mathcal{P}_g$ & $\mathcal{R}_g$ & $\mathcal{F}_g$ \\
\hline
RGB only & 52.4 & 50.1 & 51.2 \\
RGB + Text & 54.9 & 52.7 & 53.8 \\
RGB + LiDAR & 57.8 & 55.3 & 56.5 \\
RGB + LiDAR + Text & \textbf{61.2} & \textbf{58.8} & \textbf{59.3} \\
\hline
\end{tabular}
\end{table}

Table~\ref{tab:modality_ablation} demonstrates the complementary nature of RGB, LiDAR, and textual modalities in LiGAR. The RGB-only baseline achieves an $\mathcal{F}_g$ of 51.2\%, while adding LiDAR data significantly boosts performance to 56.5\%, underscoring the value of 3D structural information. The RGB-Text combination yields an $\mathcal{F}_g$ of 53.8\%, highlighting the contextual benefits of textual data. The full multi-modal configuration achieves the best performance with an $\mathcal{F}_g$ of 59.3\%, an 8.1 percentage point improvement over the RGB-only baseline. This substantial gain, coupled with consistent improvements in precision and recall, validates our hypothesis that integrating complementary modalities significantly enhances group activity recognition. The synergistic effect of combining RGB, LiDAR, and textual data in LiGAR enables a more comprehensive understanding of group activities, demonstrating the effectiveness of our multi-modal fusion approach in leveraging diverse data sources for robust recognition.

\subsubsection{Contribution of Each Component}

\begin{table}[h]
\centering
\caption{Ablation study on LiGAR components}
\label{tab:component_ablation}
\begin{tabular}{lccc}
\hline
\textbf{Model Variant} & $\mathcal{P}_g$ & $\mathcal{R}_g$ & $\mathcal{F}_g$ \\
\hline
LiGAR w/o MLT & 56.7 & 54.2 & 55.4 \\
LiGAR w/o CMGA & 58.3 & 55.9 & 57.1 \\
LiGAR w/o AFM & 59.5 & 57.1 & 58.3 \\
\textbf{Full LiGAR} & \textbf{61.2} & \textbf{58.8} & \textbf{59.3} \\
\hline
\end{tabular}
\end{table}
Table~\ref{tab:component_ablation} illustrates the critical role of each LiGAR component. The Multi-Scale LiDAR Transformer (MLT) proves most crucial, with its removal causing a 3.9 percentage point drop in $\mathcal{F}_g$ to 55.4\%, emphasizing the importance of multi-scale LiDAR processing. Removing the Cross-Modal Guided Attention (CMGA) reduces $\mathcal{F}_g$ to 57.1\%, highlighting its effectiveness in cross-modal feature alignment. The Adaptive Fusion Module (AFM) contributes a 1.0 percentage point improvement, refining final representations for more accurate predictions. Notably, the full LiGAR model's performance (59.3\% $\mathcal{F}_g$) exceeds the sum of individual component contributions, suggesting a synergistic effect. The consistent improvements in precision and recall across components indicate their effectiveness in reducing false positives and negatives. This ablation study confirms that each LiGAR component - MLT, CMGA, and AFM - is vital in creating a robust framework for multi-modal group activity recognition.

\subsubsection{Impact of Hierarchical Processing}
\begin{table}[h]
\centering
\caption{Impact of different hierarchical levels}
\label{tab:hierarchy_ablation}
\begin{tabular}{lccc}
\hline
\textbf{Hierarchical Levels} & $\mathcal{P}_g$ & $\mathcal{R}_g$ & $\mathcal{F}_g$ \\
\hline
Single-level & 55.8 & 53.4 & 54.6 \\
Two-level & 58.9 & 56.5 & 57.7 \\
Three-level (Full LiGAR) & \textbf{61.2} & \textbf{58.8} & \textbf{59.3} \\
\hline
\end{tabular}
\end{table}

Table~\ref{tab:hierarchy_ablation} reveals the significance of LiGAR's multi-scale approach. The single-level variant achieves an $\mathcal{F}_g$ of 54.6\%, while introducing a second level boosts performance to 57.7\%, a 3.1 percentage point increase. The full three-level model further improves $\mathcal{F}_g$ to 59.3\%, demonstrating the optimal balance in capturing individual actions, small group interactions, and scene-level activities. The consistent improvements in both precision and recall with increasing hierarchical levels indicate enhanced accuracy across various activity types. The 4.7 percentage point gap between single-level and three-level models underscores the crucial role of hierarchical processing in LiGAR, enabling effective leverage of information from different spatial and temporal resolutions. This study confirms that LiGAR's hierarchical processing is essential for its superior performance, with each level contributing meaningfully to the model's ability to recognize the multi-granular nature of group interactions. We performed all the above ablation studies for Volleyball and NBA datasets in \textit{Sec. 2} of supplementary material.
\subsection{Visualization}

\begin{figure*}
    \centering
    \includegraphics[width=0.7\linewidth]{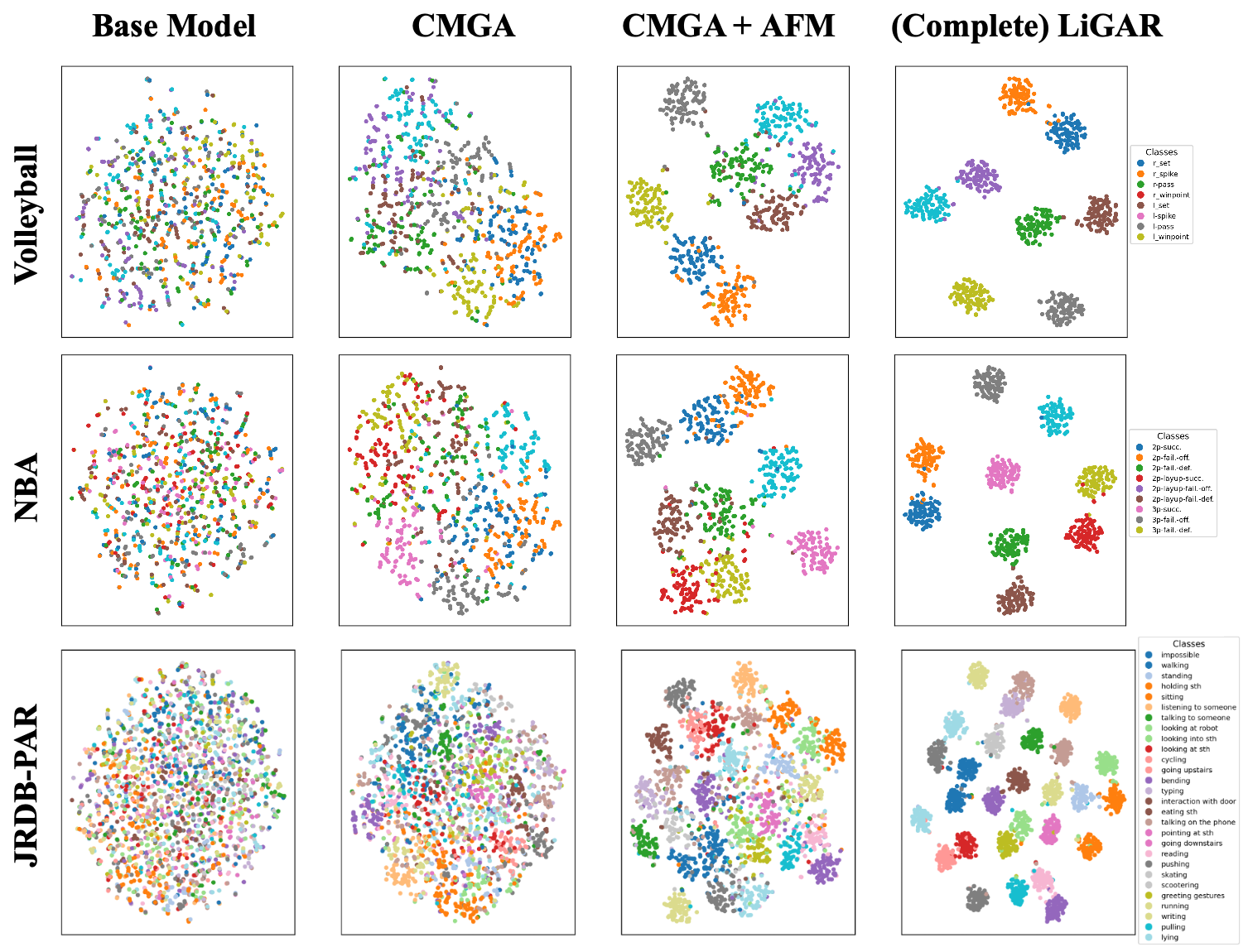}
    \caption{t-SNE~\cite{van2008visualizing} visualization of video representation on the Volleyball~\cite{ibrahim2016hierarchical}, NBA~\cite{yan2020social} and JRDB-PAR~\cite{han2022panoramic} datasets learned by different variants of our LiGAR model. \textbf{Best viewed in zoom and color.}}
    \label{fig:tsne_visualization}
\end{figure*}

Figure~\ref{fig:tsne_visualization} presents t-SNE visualizations of feature representations learned by LiGAR and its variants across the Volleyball, NBA, and JRDB-PAR datasets. The progression from the Base Model to the complete LiGAR demonstrates a clear enhancement in feature discrimination. In the Base Model, classes are largely intermixed, indicating poor separability. The introduction of CMGA initiates discernible clustering, suggesting improved cross-modal feature alignment. Adding AFM further refines these clusters, highlighting its role in adaptive feature fusion. The complete LiGAR achieves the most distinct and compact clusters across all datasets, particularly evident in the complex JRDB-PAR scenario. This visual analysis corroborates our quantitative results, illustrating LiGAR's superior ability to learn discriminative features for diverse group activities. The consistent improvement pattern across datasets underscores the generalizability and effectiveness of LiGAR's multi-modal, hierarchical approach in capturing the nuanced dynamics of group interactions. We present more t-SNE visualizations for different combinations of modalities in Sec. 3 of the supplementary material.

\section{Discussion}\label{sec:discussion}

The LiGAR framework addresses critical challenges in group activity recognition through its innovative multi-modal and hierarchical approach. At its core, LiGAR leverages \textbf{LIDAR data as a structural backbone}, enabling effective integration of visual and textual information. This multi-modal fusion, dynamically weighted based on context, enhances robustness across diverse scenarios. The framework's \textbf{multi-scale processing} aligns with the inherently hierarchical nature of group activities, capturing interactions from individual actions to scene-level dynamics. LiGAR's ability to maintain \textbf{temporal consistency} in predictions, guided by LIDAR motion information, is crucial for understanding the evolution of group activities over time. The model's \textbf{simplified loss function}, combining task-specific objectives with hierarchical and temporal consistency, provides a balanced training approach without the pitfalls of overly complex formulations. By leveraging the strengths of each modality - LIDAR's precise spatial information, video's appearance details, and text's semantic context - LiGAR demonstrates a novel approach to handling common challenges in real-world scenarios, such as varying environmental conditions and occlusions. This innovative integration of LIDAR as a guiding modality represents a significant advancement in group activity recognition, paving the way for more robust and adaptable AI systems in complex, real-world environments.

\section{Conclusion}

We introduce LiGAR, a multi-modal framework for group activity recognition that utilizes LiDAR data to enhance visual and textual information processing. LiGAR's Multi-Scale LiDAR Transformer, Cross-Modal Guided Attention, and Adaptive Fusion Module achieve state-of-the-art performance across various datasets, effectively capturing group dynamics from individual actions to scene-level interactions.
Extensive experiments validate the benefits of LiGAR's components and multi-modal approach. Notably, LiGAR maintains high accuracy even without LiDAR data during inference, demonstrating its robustness and efficiency. These advancements position LiGAR as a versatile solution with applications ranging from sports analytics to surveillance.

\noindent\textbf{Limitations:} The reliance on high-quality LiDAR data for optimal performance may restrict its applicability in resource-constrained environments. Additionally, while our model demonstrates robust performance across multiple datasets, its generalization to extremely diverse or previously unseen group activities remains to be fully explored. Future work could address these limitations by investigating techniques for operating with lower-quality LiDAR inputs and incorporating more diverse training data to enhance generalization. Exploring the integration of unsupervised learning techniques could also potentially reduce the dependence on large annotated datasets, further broadening LiGAR's applicability in real-world scenarios.

\noindent\textbf{Acknowledgments:} We would like to acknowledge Arkansas Biosciences Institute (ABI) Grant, and the NSF Data Science, Data Analytics that are Robust and Trusted (DART) for their funding in supporting this research.
{
    \small
    \bibliographystyle{ieee_fullname}
    \bibliography{main}
}

\end{document}